%% file: main.tex
\pdfoutput=1
\documentclass[10pt,twocolumn,letterpaper]{article}

\usepackage[pagenumbers]{preprint}

\input{preamble}

\definecolor{linkblue}{rgb}{0.21,0.49,0.74}
\usepackage[breaklinks,colorlinks,allcolors=linkblue]{hyperref}

\title{Predicting Body-Part Engagement for Humanoid Loco-Manipulation}

\author{%
Xinzhe Chen \quad Liangming Zhao \quad Xinyi Jiang \quad Shaozheng Zheng \quad Xue Ren\\
Qiongjie Cui \quad Zeyang Liu \quad Xingyu Chen$^{\dagger}$ \quad Xuguang Lan$^{\dagger}$\\[3pt]
{\normalsize National Key Laboratory of Human-Machine Hybrid Augmented Intelligence}\\
{\normalsize Institute of Artificial Intelligence and Robotics, Xi'an Jiaotong University}
}

\makeatletter
\def\@maketitle{%
  \newpage
  \null
  \vskip .02in
  \begin{center}%
    {\Large \bf \@title \par}%
    \vspace*{6pt}%
    {\large
      \lineskip .5em
      \begin{tabular}[t]{c}\@author\end{tabular}\par}%
    \vspace*{3pt}%
  \end{center}%
}
\makeatother

\begin{document}
\twocolumn[{%
  \renewcommand\twocolumn[1][]{#1}%
  \maketitle
  {\centering
    \includegraphics[width=\textwidth]{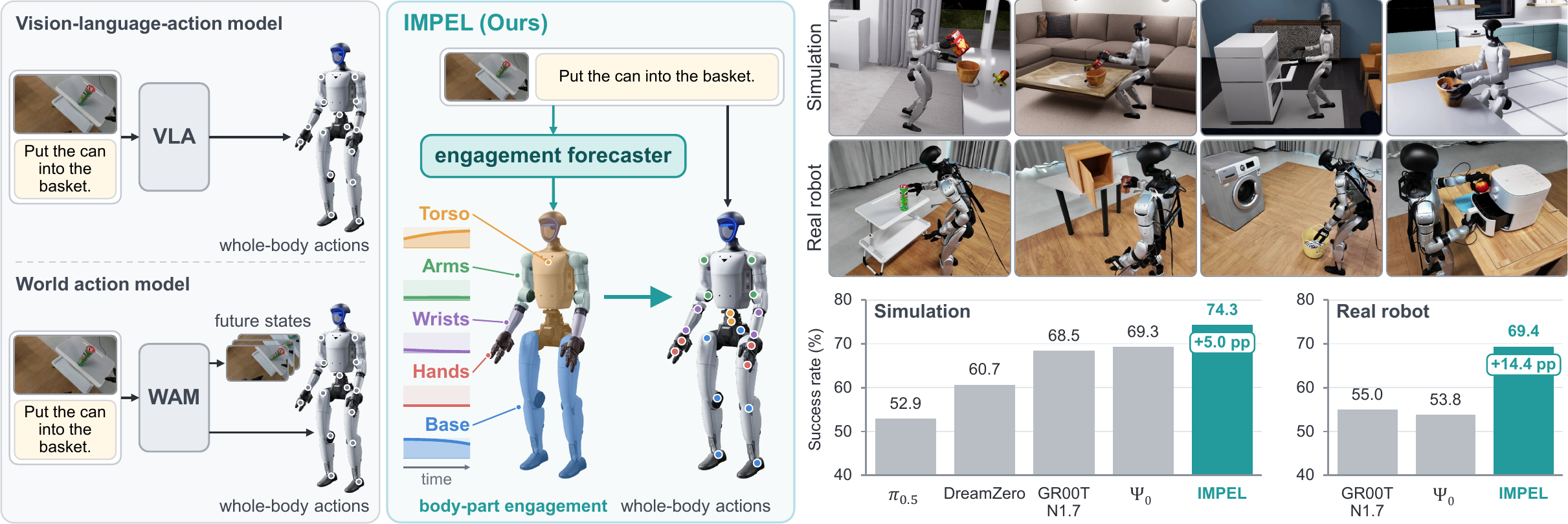}
    \captionof{figure}{Left, VLAs map images and instructions directly to whole-body actions, and WAMs also predict future states. Middle, in \method the forecaster predicts body-part engagement over the action horizon, and its latent representation conditions the action head. Right, on eight simulation and eight real-world tasks, \method achieves higher overall success than the evaluated baselines.}
    \label{fig:teaser}
  \par}\vspace{5pt}
}]
\begingroup
\renewcommand\thefootnote{}%
\begin{NoHyper}%
\footnotetext{\raggedright $^{\dagger}$Corresponding authors: \mbox{xglan@mail.xjtu.edu.cn} and \mbox{chenxingyu1990@xjtu.edu.cn}.}%
\end{NoHyper}%
\endgroup
\input{sec/0_abstract}
\input{sec/1_intro}
\input{sec/2_related}
\input{sec/3_intent}
\input{sec/3_method}
\input{sec/4_experiments}
\input{sec/6_conclusion}
{
    \small
    \bibliographystyle{ieeenat_fullname}
    \bibliography{main}
}

\end{document}

%% file: preamble.tex
\newcommand{\method}{IMPEL\xspace}

\ifdefined\pdfsuppressptexinfo \pdfsuppressptexinfo=-1 \fi

%% file: sec/0_abstract.tex
\begin{abstract}
Humanoid loco-manipulation requires the torso, arms, wrists, hands and base to coordinate their motions throughout a task, with each part serving a distinct role and following its own timing.
The prevailing approaches use vision-language-action models (VLAs) or world action models (WAMs) to generate high-dimensional whole-body actions directly from their intermediate representations.
However, this paradigm leaves the action head to implicitly infer when and how strongly each body part should engage, which can limit how well VLAs and WAMs learn loco-manipulation.
We propose \method to explicitly predict body-part engagement over the action horizon before generating whole-body actions.
Body-part engagement describes each part's target-directed participation and the hands' contact state.
For training, we recover it from teleoperation demonstrations without annotation or external models by anchoring part targets to later events.
An engagement forecaster predicts it from vision-language features and proprioception, and its latent representation conditions the action head as a prior.
We validate \method across eight simulated loco-manipulation tasks against GR00T N1.7, $\Psi_0$, $\pi_{0.5}$ and DreamZero, and eight real-world Unitree G1 tasks spanning loco-manipulation and standing manipulation.
\method reaches 74.3\% average success in simulation, 5.0 points above the strongest baseline $\Psi_0$, and 69.4\% on the real robot, 14.4 points above the best~of the evaluated baselines.
Our project page is available at \href{https://impel-humanoid.github.io/}{https://impel-humanoid.github.io/}.
\end{abstract}

%% file: sec/1_intro.tex
\section{Introduction}
\label{sec:intro}

With a human-like body that moves dynamically and dexterously, humanoids are well suited to open real-world environments, which makes humanoid loco-manipulation an increasingly important research direction in robot learning~\cite{gu2026humanoid,fu2024humanplus,he2024omnih2o}.
Current humanoid loco-manipulation models mainly follow two paradigms.
Vision-language-action models (VLAs)~\cite{bjorck2025gr00t,jiang2026wholebodyvla,wei2026psi_0} use the general spatial and semantic representations of pretrained vision-language models to generate actions.
World action models (WAMs)~\cite{yue2026being,zheng2026motionwam} jointly predict future states and actions, which gives action generation a prior informed by dynamics.
Both paradigms typically share a pipeline that collects expert data through whole-body teleoperation~\cite{ze2025twist,ben2025homie,luo2026sonic}, trains a model whose action head generates whole-body actions from observations and language instructions, and executes these actions through a whole-body controller~\cite{he2025hover,li2025amo}.

However, humanoid loco-manipulation requires coordinated motion across different body parts.
Several body parts often engage at the same time, each with its own role and timing.
Consider a humanoid picking up a can from a low table.
Its base moves the robot within reach of the table, its torso leans forward to lower the arm, and its wrist orients the hand before contact.
Which body parts engage, and how strongly, thus changes from step to step.
Existing VLAs and WAMs feed their vision-language or world-model representations directly to the action head, without explicit supervision of these engagement patterns.
The action head therefore has to implicitly infer when and how strongly each body part should engage, which can limit how well these models learn from expert demonstrations.

Even without part-level labels, expert teleoperation demonstrations reveal how each body part engages at every step through their recorded commands and subsequent task events.
We describe body-part engagement by the degree to which each part's commands drive it toward its target and, for the hands, by their contact state.
Supervising an intermediate representation to predict engagement provides whole-body action generation with structured guidance on which parts engage and how strongly at each future step.

On this basis, we propose \method (Integrating Multi-Part Engagement for Loco-manipulation) and illustrate it in \cref{fig:teaser}.
For training, we recover body-part engagement at each step from teleoperation demonstrations.
We frame the commands for torso posture, arm reaching, wrist rotation and locomotion as moving toward targets anchored to later events, such as where the hand makes contact or where the base stops.
We derive hand contact separately from the finger joints.
Recovery requires neither external models nor human annotation and applies to standard humanoid teleoperation data.
From vision-language features and the robot's proprioceptive state, an engagement forecaster predicts body-part engagement over the action horizon, and its latent representation conditions the action head as a prior.
At inference, \method generates whole-body actions from current observations without the recovery procedure.

To evaluate \method, we test eight SIMPLE loco-manipulation tasks~\cite{wei2026simple} across three randomization levels against the general-purpose VLA $\pi_{0.5}$~\cite{intelligence2025pi_}, the WAM DreamZero~\cite{ye2026world}, and the humanoid models GR00T N1.7~\cite{bjorck2025gr00t} and $\Psi_0$~\cite{wei2026psi_0}.
Across all tasks and levels, \method reaches 74.3\% average success against 68.5\% for GR00T N1.7 under the same data and training budget, and exceeds the strongest baseline $\Psi_0$ by 5.0 points.
At every randomization level, \method achieves the highest average success rate.
Across all eight simulated tasks, body-part engagement supervision gives higher average success rates than six alternative supervision variants and the same architecture trained without engagement supervision.
On a Unitree G1 humanoid, we also evaluate eight loco-manipulation and standing manipulation tasks against GR00T N1.7 and $\Psi_0$, where \method reaches 69.4\% overall success, 14.4 points above the best compared baseline.
Our contributions are:
\begin{itemize}
\item We formulate body-part engagement as structured supervision for whole-body action learning, describing target-directed participation and hand contact across \mbox{concurrently active body parts}.
\item We introduce \method to recover body-part engagement from teleoperation demonstrations without annotation or external models, and condition whole-body action generation on a latent representation trained to predict it.
  \item We demonstrate gains in simulation and on a real humanoid, with controlled ablations examining engagement supervision and its representation.
\end{itemize}

%% file: sec/2_related.tex
\section{Related Work}
\label{sec:related}

\noindent\textbf{VLAs and WAMs.}
Building on pretrained vision-language backbones, VLAs generate actions across robotic manipulation tasks~\cite{li2024cogact,qu2025spatialvla,kim2025fine,pertsch2025fast,team2025gemini}.
RT-2~\cite{brohan2023rt} and OpenVLA~\cite{kim2024openvla} emit actions as discrete language-model tokens, whereas $\pi_0$~\cite{black2024pi_0} and $\pi_{0.5}$~\cite{intelligence2025pi_} generate continuous action chunks with a flow-matching action head.
Several VLAs also produce intermediate outputs, including language motions~\cite{belkhale2024rt}, embodied reasoning~\cite{zawalski2024robotic}, subgoal images~\cite{zhao2025cot} and latent features of future observations~\cite{zheng2025flare}.
WAMs instead incorporate future observation prediction into action learning~\cite{du2023learning,hu2024video,cheang2024gr,zhu2025unified,cen2025worldvla,wu2024unleashing}, and LingBot-VA~\cite{li2026causal}, Cosmos Policy~\cite{kim2026cosmos} and DreamZero~\cite{ye2026world} jointly generate video and actions within one model.
Most of these models target single-arm, dual-arm or mobile-base platforms~\cite{10611477,khazatsky2024droid,zhao2023learning,fu2024mobile}.
Humanoid loco-manipulation instead requires coordinated commands for the torso, arms, wrists, dexterous hands and base, so extending general-purpose models to this setting requires action interfaces compatible with whole-body controllers.

\noindent\textbf{Humanoid whole-body controllers.}
Whole-body controllers map high-level commands of a humanoid to joint targets, and most are trained with reinforcement learning in simulation~\cite{makoviychuk2021isaac,rudin2022learning}.
Early learned controllers focus on bipedal locomotion~\cite{doi:10.1126/scirobotics.adi9579,zhuang2024humanoid,gu2024humanoid}, whereas motion tracking later extended such controllers to general whole-body behaviors~\cite{10377518,11696204,he2025asap,chen2025gmt,doi:10.1126/scirobotics.adx8924}.
H2O~\cite{10801984}, OmniH2O~\cite{he2024omnih2o} and ExBody~\cite{cheng2024expressive} follow retargeted human motion for teleoperation and expressive movement, and HOVER~\cite{he2025hover} unifies several command modes within one controller.
AMO~\cite{li2025amo} extends whole-body teleoperation to loco-manipulation, and SONIC~\cite{luo2026sonic} unifies walking, bending and reaching behind a latent motion-token interface driven by VR teleoperation, locomotion planners or learned models.
Collecting demonstrations through SONIC teleoperation and deploying trained models through the same interface has become a common pairing~\cite{bjorck2025gr00t,zheng2026motionwam,yue2026being}.
These controllers execute whole-body commands given by the operator during teleoperation and by a learned model at deployment.
\method keeps the controller interface unchanged and recovers body-part engagement from whole-body teleoperation demonstrations to supervise the model above the controller.

\noindent\textbf{Humanoid manipulation models.}
Humanoid manipulation models are typically learned from teleoperation demonstrations~\cite{11246340,qiu2025humanoid}.
Early systems such as Open-TeleVision~\cite{cheng2024open} and HumanPlus~\cite{fu2024humanplus} train a separate model for each task, while GR00T N1~\cite{bjorck2025gr00t} introduced an open foundation model for generalist humanoids.
Several recent models leverage intermediate motion priors from video pretraining~\cite{luo2026visionlanguageaction,yang2025egovla} or future-state prediction.
WholeBodyVLA~\cite{jiang2026wholebodyvla} learns latent actions for manipulation and locomotion from egocentric videos and decodes them into arm joint actions and locomotion commands.
Being-M0.7~\cite{yue2026being} predicts future state trajectories of the robot, and MotionWAM~\cite{zheng2026motionwam} draws on intermediate denoising features of a pretrained video world model.
GR00T N1.7~\cite{bjorck2025gr00t} and $\Psi_0$~\cite{wei2026psi_0} are recent humanoid foundation models with latent motion actions and broader pretraining, and their action heads generate whole-body actions from vision-language features.
Target-directed body-part engagement remains implicit in latent actions, predicted states, vision-language features or world-model features.
\method trains an intermediate representation to predict body-part engagement and uses this representation as a prior for \mbox{whole-body action generation}.

%% file: sec/3_intent.tex
\section{Body-Part Engagement Formulation}
\label{sec:method-intent}

\begin{figure*}[t]
  \centering
  \includegraphics[width=\textwidth]{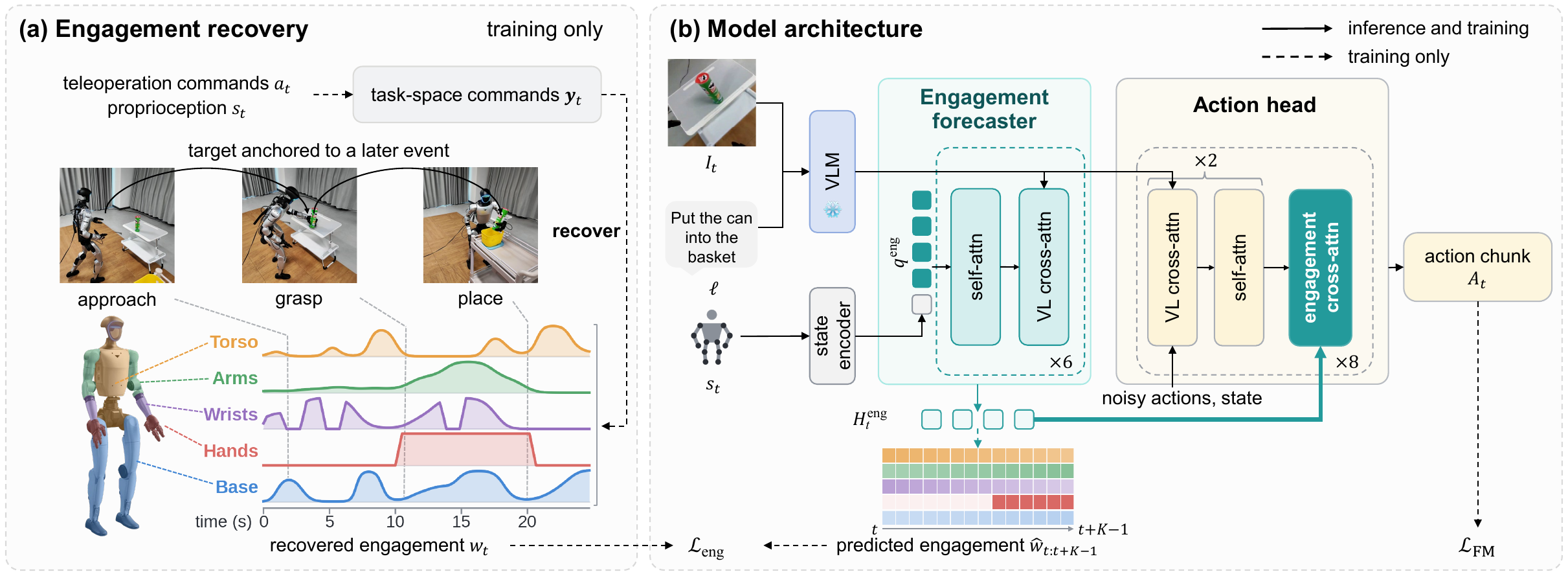}
  \caption{Overview of \method. (a) We recover body-part engagement from teleoperation demonstrations by anchoring targets to later events. (b) The forecaster outputs $H_t^{\mathrm{eng}}$, whose readout is supervised by recovered engagement. The action head attends to $H_t^{\mathrm{eng}}$ through engagement cross-attention layers, using this supervised representation to guide action generation.}
  \label{fig:pipeline}
  \vspace{-2pt}
\end{figure*}

We formalize body-part engagement for learning from whole-body teleoperation demonstrations.
At step $t$, the model observes an egocentric image $I_t$, a language instruction $\ell$ and the proprioceptive state $s_t$.
It outputs a $K$-step action chunk~\cite{zhao2023learning} $A_t = (a_t, \dots, a_{t+K-1})$, and a whole-body controller maps each $a_t$ to joint commands.

A whole-body action $a_t$ commands all body parts at once and leaves the role of each implicit.
We describe these roles operationally through body-part engagement over five body parts.
Torso adjusts the body posture, Arms position the hands, Wrists orient them, Hands contact and release objects, and Base moves the robot.
Four of these body parts, $k \in \mathcal{K} = \{\mathrm{Torso}, \mathrm{Arms}, \mathrm{Wrists}, \mathrm{Base}\}$, each have a target, namely a torso posture, a wrist position, a wrist orientation, and a robot position and heading, respectively.
The target-directed engagement of body part $k$ at step $t$ is $w_t^k \in [0,1]$, the degree to which its commands drive it toward that target.
For Hands, we describe engagement by the contact state $c_t \in [0,1]$, which indicates whether a hand is in contact with an object.
Together, they form the five-channel body-part engagement $\mathbf{w}_t = (w_t^{\mathrm{Torso}}, w_t^{\mathrm{Arms}}, w_t^{\mathrm{Wrists}}, c_t, w_t^{\mathrm{Base}}) \in [0,1]^5$.
Arms, Wrists and Hands each cover the left and right sides with one channel.
The channels need not sum to one, as several parts can engage simultaneously.

A small command can reflect either low engagement or proximity to the target, so we adapt a mixture-of-controllers model~\cite{chericoni2026neural} for the four parts in $\mathcal{K}$.
The task-space command $\mathbf{y}_t$ stacks the torso posture rate, both wrists' linear and angular velocities and the base velocity, and follows
\begin{equation}
  \mathbf{y}_t = \sum_{k \in \mathcal{K}} w_t^k \, \mathbf{g}^k \odot \mathbf{e}_t^k + \boldsymbol{\epsilon}_t ,
  \label{eq:command-model}
\end{equation}
where $\mathbf{e}_t^k$ is a target-directed field constructed from the displacement of body part $k$ to its target, $\odot$ is the elementwise product and $\boldsymbol{\epsilon}_t$ is Gaussian noise.
The gains $\mathbf{g}^k$ are shared across demonstrations of a task, and each term is zero outside the task-space coordinates of its body part.
Each term models a target-directed command scaled by $w_t^k$, so at $w_t^k = 1$ it equals $\mathbf{g}^k \odot \mathbf{e}_t^k$ and at $w_t^k = 0$ it vanishes.

%% file: sec/3_method.tex
\section{Method}
\label{sec:method}

\method predicts body-part engagement before generating whole-body actions.
\Cref{fig:pipeline} shows its two parts.
We recover body-part engagement from teleoperation demonstrations to build training targets.
An engagement forecaster predicts body-part engagement from the observation. Its latent representation conditions the action head.
At inference, the model generates whole-body actions from current observations without recovery.

\subsection{Engagement Recovery}
\label{sec:method-engine}

Recovering the engagements in \cref{eq:command-model} requires targets that demonstrations leave implicit.
We anchor them to later events, with target state $\mathbf{x}_{t^\star}^k$ at step $t^\star>t$.
Arms and Wrists anchor to the next commanded hand contact, detected as a sustained rise of the finger command, and while the hand holds an object, to waypoints from the simplified wrist path.
A hand without commanded contact gives its wrist no target.
Torso anchors to the next commanded contact or release, and Base to the end of its current motion segment, a run of nonzero base command.

With these anchored targets, we compute command velocities $\mathbf{y}_t^k$ and displacements $\mathbf{d}_t^k$ for each body part $k$ from the recorded commands and robot states as
\begin{equation}
  \mathbf{y}_t^k = \frac{\tilde{\mathbf{x}}_{t+1}^k \ominus \tilde{\mathbf{x}}_t^k}{\Delta t} , \qquad
  \mathbf{d}_t^k = \mathbf{x}_{t^\star}^k \ominus \mathbf{x}_t^k ,
  \label{eq:command-displacement}
\end{equation}
where $\tilde{\mathbf{x}}_t^k$ is the commanded state and $\mathbf{x}_t^k$ the current state, measured for Arms and Wrists and commanded for Torso and Base. Wrist poses come from forward kinematics.
Here $\Delta t$ is the time step and $\ominus$ takes position or rotation differences.
During a reach, $\mathbf{d}_t^{\mathrm{Arms}}$ is the remaining displacement to the wrist position at the next commanded contact.
From $\mathbf{d}_t^k$, we construct the field $\mathbf{e}_t^k$ in \cref{eq:command-model} with saturation and Arms target-motion compensation, and apply common channelwise scaling to $\mathbf{y}_t^k$ and $\mathbf{e}_t^k$.

We fit \cref{eq:command-model} with smooth engagements $w_t^k=\operatorname{sigmoid}(\boldsymbol{\phi}_t(\eta)^\top\boldsymbol{\beta}^k)$, where $\boldsymbol{\phi}_t(\eta)$ contains Gaussian basis functions of width $\eta$ and $\boldsymbol{\beta}^k$ their coefficients~\cite{chericoni2026neural}.
We fix task gains $\mathbf{g}^k$ estimated from 20 demonstrations and fit $\boldsymbol{\theta}=(\boldsymbol{\beta},\eta,\sigma)$ to each demonstration, with noise scale $\sigma$,
\begin{equation}
  \begin{aligned}
    \hat{\boldsymbol{\theta}} = \arg\min_{\boldsymbol{\theta}} \;
      & \frac{1}{2\sigma^2} \sum_t \Big\| \mathbf{y}_t - \sum_{k \in \mathcal{K}} w_t^k \, \mathbf{g}^k \odot \mathbf{e}_t^k \Big\|^2 \\
      & + n\log\sigma + \mathcal{R}(\boldsymbol{\theta}) ,
  \end{aligned}
  \label{eq:engine-map}
\end{equation}
where $n$ counts scalar observations and $\mathcal{R}$ collects engagement regularizers and parameter priors.
The scale prior pulls each active segment's peak engagement toward one.
For Arms, the fit gives high engagement when the wrist velocity command is close to $\mathbf{g}^{\mathrm{Arms}}\odot\mathbf{e}_t^{\mathrm{Arms}}$, and low engagement when the command is near zero despite a remaining displacement to the anchored target.

\input{sec/table_main}

The fitted $\hat{\boldsymbol{\beta}}^k$ and $\hat\eta$ give the four target-directed engagements. Finger joints determine binary Hands contact $b_t$,
\begin{align}
  w_t^k &= \operatorname{sigmoid} \big( \boldsymbol{\phi}_t(\hat\eta)^{\top} \hat{\boldsymbol{\beta}}^k \big) , \quad k \in \mathcal{K} , \label{eq:engine-output} \\
  b_t &= \max_h \, \mathbb{I} \big[ \|\tilde{\mathbf{f}}_t^h\| > \kappa \, \wedge \, r_t^h < \rho \big] , \label{eq:engine-contact}
\end{align}
where $\tilde{\mathbf{f}}_t^h$ and $\mathbf{f}_t^h$ are the commanded and measured finger joints of hand $h$, $r_t^h$ is the median of $\|\mathbf{f}_s^h\| / \|\tilde{\mathbf{f}}_s^h\|$ on its closing interval, and $\kappa$ and $\rho$ are thresholds.
Thus $b_t = 1$ when an object blocks the closing fingers, and a 0.5\,s moving average smooths $b_t$ into $c_t \in [0,1]$.
Together they form $\mathbf{w}_t = (w_t^{\mathrm{Torso}}, w_t^{\mathrm{Arms}}, w_t^{\mathrm{Wrists}}, c_t, w_t^{\mathrm{Base}})$.
We also recover an activity mask $m_t^k \in [0,1]$, marking the active steps where body part $k$ pursues a target, and the uncertainty $u_t^k$ of $w_t^k$ under a Laplace approximation of the fit.
The procedure uses the same settings for all sixteen simulated and real-world tasks, spanning two whole-body controllers, and requires no object poses, annotation or external models.

\subsection{Engagement Forecaster}
\label{sec:method-forecast}

The engagement forecaster predicts body-part engagement for the $K$ steps of the action chunk.
Its inputs are the proprioceptive state $s_t$ and vision-language features extracted from $I_t$ and $\ell$.
The forecaster is a transformer~\cite{vaswani2017attention} over $K+1$ tokens.
One token embeds $s_t$.
The $K$ learned queries $q^{\mathrm{eng}}$ each represent one chunk step and receive sinusoidal step embeddings.
Every layer applies self-attention among all tokens without a causal mask, followed by cross-attention to the vision-language features that the action head also attends to.
The outputs at the $K$ queries form the engagement representation $H_t^{\mathrm{eng}} \in \mathbb{R}^{K \times d}$, where $d$ is the hidden size.
A linear layer with a sigmoid maps each row $i$ of $H_t^{\mathrm{eng}}$ to the predicted engagement $\hat{\mathbf{w}}_{t+i} \in [0,1]^5$.

Over the $K$ steps, the engagement loss compares the readout $\hat{\mathbf{w}}_{t+i}$ with the recovered $\mathbf{w}_{t+i}$,
\begin{equation}
  \mathcal{L}_{\mathrm{eng}} = \tfrac{1}{Z} {\textstyle \sum_{i=0}^{K-1}} \boldsymbol{\omega}_{t+i}^{\top} \, \mathrm{BCE}\big(\hat{\mathbf{w}}_{t+i}, \mathbf{w}_{t+i}\big) ,
  \label{eq:intent-loss}
\end{equation}
where $\mathrm{BCE}$ is the channelwise binary cross-entropy and $Z$ is the sum of the weights.
For the four target-directed channels, the entry of $\boldsymbol{\omega}_{t+i}$ is $m_{t+i}^k \exp(-u_{t+i}^k / \tau)$, where $\tau$ is a temperature.
The Hands channel has weight one.
These weights restrict supervision of the four target-directed channels to active intervals and downweight uncertain estimates.

\subsection{Engagement-Conditioned Action Generation}
\label{sec:method-action}

The action head generates actions conditioned on $H_t^{\mathrm{eng}}$ by iterative denoising from Gaussian noise.
At each denoising step, the noisy actions and the state $s_t$ enter as tokens and pass through $L$ units.
Each unit repeats vision-language cross-attention followed by self-attention twice, then applies engagement cross-attention.
In this layer, the tokens attend to $H_t^{\mathrm{eng}}$ rather than its readout $\hat{\mathbf{w}}$.
We add sinusoidal embeddings of the chunk steps to the keys and zero-initialize the attention output projection.
The last denoising step outputs the action chunk $A_t$.
We train the model with
\begin{equation}
  \mathcal{L} = \lambda \cdot \mathcal{L}_{\mathrm{eng}} + \mathcal{L}_{\mathrm{FM}} ,
  \label{eq:loss}
\end{equation}
where $\lambda$ weights the engagement loss and $\mathcal{L}_{\mathrm{FM}}$ is the flow-matching loss~\cite{lipman2023flow} on the action chunk $A_t$.
The gradient of $\mathcal{L}_{\mathrm{FM}}$ also reaches the forecaster through $H_t^{\mathrm{eng}}$, so the engagement representation is shaped by both $\mathcal{L}_{\mathrm{eng}}$ and $\mathcal{L}_{\mathrm{FM}}$.

%% file: sec/table_main.tex
\begin{table*}[t]
  \caption{Success rate (\%) on eight SIMPLE tasks with the official scenes and success criterion. Each model runs three seeds, each covering 10 scenes per task at each of three randomization levels (DR0, DR1, DR2), 80 episodes per level and 240 overall. Cells give the mean and standard deviation over seeds, and bold marks the highest mean in each column. The lower block has the same model rows.}
  \label{tab:main}
  \centering
  \newcommand{\sd}[1]{{\hspace{1pt}\scriptsize\color{black!50}\textpm#1}}
  \footnotesize
  \renewcommand{\arraystretch}{1.2}
  \setlength{\tabcolsep}{0pt}
  \scalebox{0.94}{\begin{tabular*}{\dimexpr\textwidth*100/94\relax}{@{\extracolsep{\fill}}l*{12}{r@{\extracolsep{0pt}}l@{\extracolsep{\fill}}}r@{\extracolsep{0pt}}l@{}}
    \toprule
     & \multicolumn{8}{c}{All tasks} & \multicolumn{6}{c}{LocomotionPickBetweenTables} & \multicolumn{6}{c}{XMovePick} & \multicolumn{6}{c}{XMoveBendPick} \\
    \cmidrule(lr){2-9} \cmidrule(lr){10-15} \cmidrule(lr){16-21} \cmidrule(lr){22-27}
    Model & \multicolumn{2}{c}{Avg} & \multicolumn{2}{c}{DR0} & \multicolumn{2}{c}{DR1} & \multicolumn{2}{c}{DR2} & \multicolumn{2}{c}{DR0} & \multicolumn{2}{c}{DR1} & \multicolumn{2}{c}{DR2} & \multicolumn{2}{c}{DR0} & \multicolumn{2}{c}{DR1} & \multicolumn{2}{c}{DR2} & \multicolumn{2}{c}{DR0} & \multicolumn{2}{c}{DR1} & \multicolumn{2}{c}{DR2} \\
    \midrule
    $\pi_{0.5}$~\cite{intelligence2025pi_} & 52.9 & \sd{2.1} & 58.3 & \sd{5.1} & 57.5 & \sd{1.3} & 42.9 & \sd{1.9} & 43.3 & \sd{25.2} & 26.7 & \sd{5.8} & 6.7 & \sd{5.8} & \textbf{100.0} & \sd{0.0} & \textbf{100.0} & \sd{0.0} & 70.0 & \sd{10.0} & 0.0 & \sd{0.0} & 0.0 & \sd{0.0} & 10.0 & \sd{0.0} \\
    DreamZero~\cite{ye2026world} & 60.7 & \sd{0.9} & 63.3 & \sd{0.7} & 66.7 & \sd{6.2} & 52.1 & \sd{3.1} & 30.0 & \sd{17.3} & 33.3 & \sd{28.9} & 10.0 & \sd{0.0} & 86.7 & \sd{5.8} & 93.3 & \sd{5.8} & 76.7 & \sd{5.8} & 36.7 & \sd{25.2} & 26.7 & \sd{11.5} & 33.3 & \sd{11.5} \\
    GR00T N1.7~\cite{bjorck2025gr00t} & 68.5 & \sd{3.4} & 75.8 & \sd{1.4} & 71.3 & \sd{1.3} & 58.3 & \sd{8.9} & 56.7 & \sd{5.8} & 53.3 & \sd{20.8} & 10.0 & \sd{0.0} & 93.3 & \sd{5.8} & 76.7 & \sd{5.8} & \textbf{83.3} & \sd{5.8} & 80.0 & \sd{0.0} & 63.3 & \sd{5.8} & 83.3 & \sd{5.8} \\
    $\Psi_0$~\cite{wei2026psi_0} & 69.3 & \sd{2.3} & 80.4 & \sd{3.6} & 73.8 & \sd{8.8} & 53.8 & \sd{1.3} & 53.3 & \sd{11.5} & 56.7 & \sd{25.2} & \textbf{30.0} & \sd{0.0} & \textbf{100.0} & \sd{0.0} & 96.7 & \sd{5.8} & 60.0 & \sd{0.0} & \textbf{90.0} & \sd{0.0} & \textbf{73.3} & \sd{20.8} & 70.0 & \sd{10.0} \\
    \midrule
    \method{} (ours) & \textbf{74.3} & \sd{2.1} & \textbf{82.9} & \sd{4.4} & \textbf{75.8} & \sd{1.4} & \textbf{64.2} & \sd{1.9} & \textbf{63.3} & \sd{15.3} & \textbf{60.0} & \sd{10.0} & 20.0 & \sd{10.0} & \textbf{100.0} & \sd{0.0} & 90.0 & \sd{10.0} & 73.3 & \sd{5.8} & 86.7 & \sd{5.8} & \textbf{73.3} & \sd{11.5} & \textbf{86.7} & \sd{5.8} \\
    \midrule[\heavyrulewidth]
  \end{tabular*}}
  \par\nointerlineskip
  \scalebox{0.94}{\begin{tabular*}{\dimexpr\textwidth*100/94\relax}{@{\extracolsep{\fill}}*{14}{r@{\extracolsep{0pt}}l@{\extracolsep{\fill}}}r@{\extracolsep{0pt}}l@{}}
    \addlinespace[1pt]
    \multicolumn{6}{c}{BendPickAndPlace} & \multicolumn{6}{c}{BendHandover} & \multicolumn{6}{c}{PickAndPlaceAndHugContainer} & \multicolumn{6}{c}{OpenOven} & \multicolumn{6}{c}{PushOfficeChair} \\
    \cmidrule(lr){1-6} \cmidrule(lr){7-12} \cmidrule(lr){13-18} \cmidrule(lr){19-24} \cmidrule(lr){25-30}
    \multicolumn{2}{c}{DR0} & \multicolumn{2}{c}{DR1} & \multicolumn{2}{c}{DR2} & \multicolumn{2}{c}{DR0} & \multicolumn{2}{c}{DR1} & \multicolumn{2}{c}{DR2} & \multicolumn{2}{c}{DR0} & \multicolumn{2}{c}{DR1} & \multicolumn{2}{c}{DR2} & \multicolumn{2}{c}{DR0} & \multicolumn{2}{c}{DR1} & \multicolumn{2}{c}{DR2} & \multicolumn{2}{c}{DR0} & \multicolumn{2}{c}{DR1} & \multicolumn{2}{c}{DR2} \\
    \midrule
    33.3 & \sd{11.5} & 40.0 & \sd{17.3} & 30.0 & \sd{20.0} & 86.7 & \sd{5.8} & 86.7 & \sd{5.8} & \textbf{93.3} & \sd{5.8} & 63.3 & \sd{5.8} & \textbf{66.7} & \sd{5.8} & \textbf{53.3} & \sd{11.5} & 40.0 & \sd{20.0} & 40.0 & \sd{10.0} & 16.7 & \sd{15.3} & \textbf{100.0} & \sd{0.0} & \textbf{100.0} & \sd{0.0} & 63.3 & \sd{11.5} \\
    50.0 & \sd{10.0} & \textbf{73.3} & \sd{11.5} & 56.7 & \sd{5.8} & 80.0 & \sd{10.0} & 70.0 & \sd{10.0} & 66.7 & \sd{20.8} & 33.3 & \sd{11.5} & 50.0 & \sd{17.3} & 33.3 & \sd{20.8} & \textbf{93.3} & \sd{11.5} & \textbf{90.0} & \sd{10.0} & \textbf{90.0} & \sd{10.0} & 96.7 & \sd{5.8} & 96.7 & \sd{5.8} & 50.0 & \sd{0.0} \\
    43.3 & \sd{15.3} & 60.0 & \sd{0.0} & 50.0 & \sd{20.0} & 63.3 & \sd{5.8} & 76.7 & \sd{5.8} & 63.3 & \sd{20.8} & \textbf{76.7} & \sd{5.8} & 63.3 & \sd{5.8} & 43.3 & \sd{5.8} & \textbf{93.3} & \sd{5.8} & 76.7 & \sd{20.8} & 63.3 & \sd{20.8} & \textbf{100.0} & \sd{0.0} & \textbf{100.0} & \sd{0.0} & \textbf{70.0} & \sd{0.0} \\
    63.3 & \sd{5.8} & 50.0 & \sd{0.0} & 46.7 & \sd{5.8} & \textbf{93.3} & \sd{5.8} & 86.7 & \sd{5.8} & 83.3 & \sd{5.8} & 66.7 & \sd{15.3} & 50.0 & \sd{10.0} & 33.3 & \sd{15.3} & 76.7 & \sd{25.2} & 76.7 & \sd{20.8} & 36.7 & \sd{15.3} & \textbf{100.0} & \sd{0.0} & \textbf{100.0} & \sd{0.0} & \textbf{70.0} & \sd{0.0} \\
    \midrule
    \textbf{70.0} & \sd{10.0} & 60.0 & \sd{17.3} & \textbf{63.3} & \sd{11.5} & 86.7 & \sd{5.8} & \textbf{96.7} & \sd{5.8} & 76.7 & \sd{11.5} & 73.3 & \sd{5.8} & 53.3 & \sd{11.5} & 43.3 & \sd{11.5} & 86.7 & \sd{5.8} & 73.3 & \sd{11.5} & 80.0 & \sd{10.0} & 96.7 & \sd{5.8} & \textbf{100.0} & \sd{0.0} & \textbf{70.0} & \sd{0.0} \\
    \bottomrule
  \end{tabular*}}
\end{table*}

%% file: sec/4_experiments.tex
\section{Experiments}
\label{sec:exp}

We answer four research questions: Q1. Does \method achieve higher success than VLA and WAM baselines on humanoid loco-manipulation? Q2. How do engagement supervision and other design choices contribute? Q3. How accurately does \method predict engagement, and how accurately is it recovered from commands? Q4. Does the advantage carry over to a real humanoid?

\subsection{Implementation Details}
\label{sec:exp-setup}

We build \method on GR00T N1.7~\cite{bjorck2025gr00t} and keep its vision-language model frozen.
The engagement forecaster attends to the same vision-language features as the GR00T N1.7 action head, and its $K = 40$ queries match the 40-step action chunk, 0.8\,s at 50\,Hz.
It has 6 layers, 24 heads and a width of $d = 1536$, equal to that of the action head.
We insert the $L = 8$ engagement cross-attention layers after every fourth block of the 32-block action head.
The forecaster adds 230M parameters and the cross-attention layers add 76M, both trained from scratch, and the action head is fine-tuned from GR00T N1.7.
We train \method per task for 40k steps at batch size 32 on one A100 GPU, using AdamW~\cite{loshchilov2018decoupled} with learning rate $10^{-4}$ and loss weight $\lambda = 0.1$.

\begin{figure*}[t]
  \begin{minipage}[t]{\columnwidth}
    \vspace{0pt}\centering
    \includegraphics[width=\linewidth]{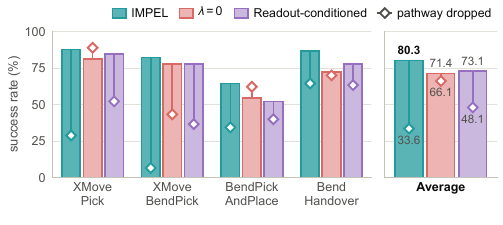}
  \end{minipage}\hfill
  \begin{minipage}[t]{\columnwidth}
    \vspace{0pt}\centering
    \includegraphics[width=\linewidth]{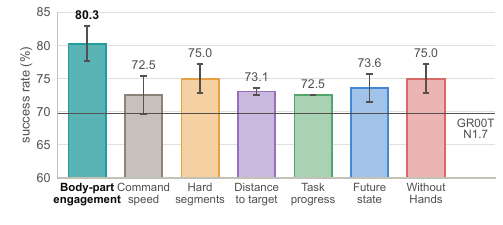}
  \end{minipage}\\
  \begin{minipage}[b]{\columnwidth}
    \caption{Success rate (\%) on each ablation task and on average, mean over three seeds. Bars keep the engagement pathway and diamonds mark the same model with it dropped at inference.}
    \label{fig:pathway}
  \end{minipage}\hfill
  \begin{minipage}[b]{\columnwidth}
    \caption{Success rate (\%) averaged over the four ablation tasks for each forecaster supervision variant, mean and standard deviation over three seeds. Teal marks body-part engagement.}
    \label{fig:supervision}
  \end{minipage}
  \vspace{-18pt}
\end{figure*}

\subsection{Simulation Experiments}
\label{sec:exp-main}

\noindent\textbf{Simulation benchmark.}
We use eight loco-manipulation tasks of the SIMPLE benchmark~\cite{wei2026simple}, which simulates a Unitree G1 with Dex3 hands, with the official scenes and success criterion.
Each task provides about 100 teleoperation demonstrations for training and has 10 scenes at each of three randomization levels, DR0, DR1 and DR2.
DR0 places the robot, the target object and the container as in the training demonstrations and randomizes the table material and distractors.
DR1 further changes the lighting.
DR2, the most randomized level, also perturbs object positions.
Each model runs the same 30 scenes per task, and we report the mean and standard deviation over three evaluation seeds.

\noindent\textbf{Baselines.}
Our matched baseline is GR00T N1.7~\cite{bjorck2025gr00t} fine-tuned with the same data and training budget as \method, with its vision-language model also frozen.
We also fine-tune the VLA $\pi_{0.5}$~\cite{intelligence2025pi_} and the WAM DreamZero~\cite{ye2026world} for each task on the same SIMPLE training data as \method for 40k steps, following the configurations released for SIMPLE.
$\Psi_0$~\cite{wei2026psi_0}, a humanoid manipulation model officially supported by SIMPLE, is fine-tuned for each task for 40k steps with its official training configuration.

\noindent\textbf{Main results.}
As \cref{tab:main} shows, \method reaches the highest average success rate, 74.3\%, against 69.3\% for $\Psi_0$, 68.5\% for GR00T N1.7, 60.7\% for DreamZero and 52.9\% for $\pi_{0.5}$.
Against the matched GR00T N1.7, \method gains 5.8 points, led by BendHandover and BendPickAndPlace.
\method also has the highest mean at every randomization level.
At DR2, it leads GR00T N1.7 by 5.9 points and $\Psi_0$ by 10.4 points.

\subsection{Ablation Studies}
\label{sec:exp-ablation}

To verify the contribution of body-part engagement supervision and other design choices, we ablate \method on the only three tasks that engage Torso, namely BendHandover, BendPickAndPlace and XMoveBendPick, and on XMovePick, the counterpart of XMoveBendPick without Torso engagement, and report their average success rate.

\begin{figure}[t]
  \centering
  \includegraphics[width=\linewidth]{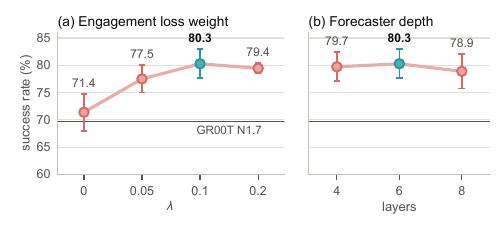}
  \caption{Success rate (\%) averaged over the four ablation tasks for (a) the engagement loss weight $\lambda$ and (b) the forecaster depth, mean and standard deviation over three seeds. Teal marks \method.}
  \label{fig:ablation}
  \vspace{-12pt}
\end{figure}

\noindent\textbf{Engagement supervision and pathway.}
To separate the roles of engagement supervision and the engagement pathway, we train the architecture of \method with $\lambda = 0$, and drop the pathway, the eight engagement cross-attention layers, from both models at inference.
On the four ablation tasks of \cref{fig:pathway}, \method reaches 80.3\% and the $\lambda = 0$ variant 71.4\%, close to GR00T N1.7 in \cref{fig:supervision,fig:ablation}.
Over all eight tasks, \method leads this variant, 74.3\% versus 69.4\%, so added capacity alone does not explain the gain.
At inference, \method relies on the engagement representation, dropping to 33.6\% without it, lower on all four tasks and most on XMoveBendPick and XMovePick, and to 38.6\% on all eight.
The $\lambda = 0$ variant drops only to 66.1\% and 59.2\%, so engagement supervision makes the action head rely on the pathway far more.
We also train a readout-conditioned variant, whose cross-attention layers attend to a linear projection of $\hat{\mathbf{w}}$ instead of $H_t^{\mathrm{eng}}$.
It reaches 73.1\% on four tasks and 71.1\% on all eight, lower than \method, and drops to 48.1\% and 50.7\% without the pathway.
Its action head uses predicted engagement, and the representation gives higher average success than the five-dimensional readout.

\begin{figure*}[t]
  \centering
  \includegraphics[width=\textwidth]{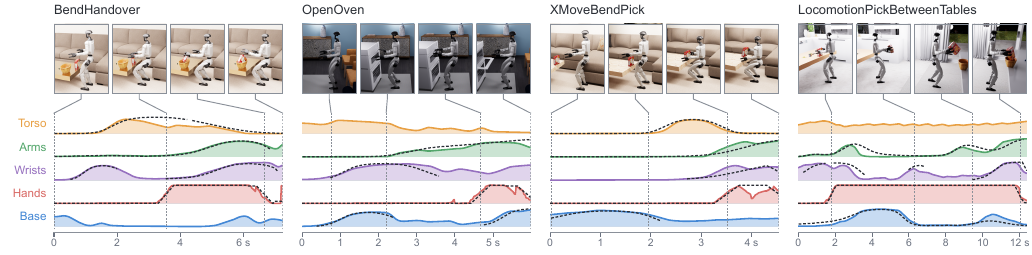}
  \caption{Engagement readout $\hat{\mathbf{w}}$ of \method on four loco-manipulation tasks. Dashed lines show engagement recovered from the executed commands, on steps where each body part is active. Thin lines link keyframes to their times.}
  \label{fig:readout}
\end{figure*}

\begin{table*}[t]
  \begin{minipage}[t]{\columnwidth}
    \input{sec/table_intent_stats}
  \end{minipage}\hfill
  \begin{minipage}[t]{\columnwidth}
    \input{sec/table_semisynth}
  \end{minipage}
\end{table*}

\noindent\textbf{Supervision targets.}
To compare supervision for the forecaster, we keep the forecaster and conditioning architecture and pair each target with its prediction head, loss and weighting.
\emph{Command speed} replaces the four target-directed channels with temporally smoothed and normalized command magnitudes and retains Hands contact.
\emph{Hard segments} threshold each of the four target-directed engagements at 0.5, \emph{distance to target} is the remaining error of each body part to its anchored target, \emph{task progress} covers episode progress, stage progress and proximity to the next contact or release, \emph{future state} is the next 40 proprioceptive states, and \emph{without Hands} removes Hands supervision.
In \cref{fig:supervision}, body-part engagement reaches 80.3\%, against 72.5\% for \emph{command speed} and 72.5 to 75.0\% for the other targets.
Averaged over all eight tasks, it reaches 74.3\%, against 68.1 to 70.7\% across the six alternatives.
Body-part engagement relates commands to part targets and includes hand contact.
With activity and uncertainty weighting, it provides more effective supervision than these alternatives.

\noindent\textbf{Engagement loss weight and forecaster depth.}
To examine the choice of $\lambda = 0.1$ and a six-layer forecaster, we retrain \method with $\lambda$ set to 0.05 and 0.2 and with forecasters of 4 and 8 layers.
As \cref{fig:ablation} shows, success is 77.5\% and 79.4\% with $\lambda$ set to 0.05 and 0.2, and 79.7\% and 78.9\% with 4 and 8 layers, so \method is not sensitive to the engagement loss weight or the forecaster depth in these ranges.

\subsection{In-Depth Analysis}
\label{sec:exp-analysis}

\noindent\textbf{Engagement visualization and statistical analysis.}
To assess whether predicted engagement follows task stages and the executed actions, \cref{fig:readout} compares the readout $\hat{\mathbf{w}}$ with engagement recovered from executed commands on four loco-manipulation tasks.
The readout follows the stages of each task.
In BendHandover, Torso rises before Arms. In OpenOven, Base rises again after contact while Arms remains engaged.
In XMoveBendPick, Base is high while the robot walks to the table and drops when it stops, then Torso rises as it bends down, Wrists and Arms rise as the hand reaches the box, and Hands switches on as the hand makes contact.
In LocomotionPickBetweenTables, Base rises only after contact, when the robot carries the box to the other table, where Arms rises and Hands switches off as it places the box.
The readout largely coincides with the recovered engagement, consistent with the executed actions.

Moreover, over all 720 evaluation episodes of \method, we correlate $\hat{\mathbf{w}}$ with engagement recovered from the executed commands, on the active steps identified during recovery.
As \cref{tab:intent-stats} shows, the pooled median is 0.92, with body-part medians from 0.75 for Wrists to 0.98 for Hands.
Against simulator hand-object contact, an independent reference, Hands reaches 0.92 for the readout and 0.96 for recovered contact.
The predictions thus track recovered engagement and physical hand contact.

\input{sec/figure_real_results}

\noindent\textbf{Recovery on semi-synthetic demonstrations.}
Since teleoperation demonstrations lack ground-truth target-directed engagement, we evaluate recovery accuracy and temporal smoothing on semi-synthetic demonstrations with known engagement curves.
For 260 teleoperation demonstrations, we use the fitted engagement curves as gradual truth and threshold them at 0.5 to obtain abrupt truth.
From each truth, we synthesize commands with \cref{eq:command-model}, using the task gains and the displacements of the demonstration and adding its real residual, shifted in time, as noise.
We then refit these commands with the procedure that produces our training targets, keeping the gains fixed.
As \cref{tab:semisynth} shows, the gradual truth is recovered with a median correlation of 0.98 and a median ramp of 1.05\,s against the true 1.00\,s.
For abrupt truth, the median correlation is 0.92, while the median ramp increases from 0.02\,s to 0.56\,s.
These results support recovery of gradual engagement curves and quantify smoothing of abrupt switches under the command model.

\subsection{Real-World Experiments}
\label{sec:exp-real}

\noindent\textbf{Setup.}
We deploy \method, its $\lambda = 0$ variant of \cref{sec:exp-ablation}, GR00T N1.7 and $\Psi_0$, the strongest baseline in simulation, on a Unitree G1 humanoid with Dex3 hands and the SONIC whole-body controller~\cite{luo2026sonic}, with SONIC motion tokens and hand joint targets as actions.
For training, we recover engagement from the recorded teleoperation stream and robot states.
As \cref{fig:real-tasks} shows, we evaluate five loco-manipulation and three standing manipulation tasks.
Each task provides about 50 real-robot teleoperation demonstrations, on which every model is trained for 40k steps.
GR00T N1.7 and both \method models use the simulation training settings of \cref{sec:exp-setup} without tuning on the real robot, and $\Psi_0$ follows its official fine-tuning recipe.
Each model runs 20 trials per task, alternating with the other three under the same lighting.
The staff who reset the scene and judge each trial are blind to which model runs.
The robot starts at a random position within about 20\,cm $\times$ 20\,cm, and the objects are at fixed positions. We also test L2, L3 and L4 with the cart, the laundry basket and the fruit basket with its cart placed at random within 30\,cm $\times$ 30\,cm.
A trial succeeds if the model completes the whole task without human intervention, and it may retry on its own.
A trial fails once the scene cannot be recovered, as when fruit spills or clothes fall, or after no progress for twice the usual completion time.

\noindent\textbf{Results.}
As \cref{tab:real} shows, \method reaches 69.4\% overall success, against 55.0\% for GR00T N1.7 and 53.8\% for $\Psi_0$, and the highest success on seven of the eight tasks.
It reaches 72.0\% on the loco-manipulation tasks and 65.0\% on the standing manipulation tasks, against at most 51.0\% and 61.7\% for the baselines.
With the object placed at random, it still leads on L2, L3 and L4, with 45.0\%, 50.0\% and 85.0\%, against at most 35.0\%, 30.0\% and 60.0\% for the other models.
The $\lambda = 0$ variant reaches 53.1\%, on par with the baselines, so the real-robot gain also depends on engagement supervision.
The eight tasks need several body parts to act together, as in L3, where \method crouches to lift the laundry basket with both hands, walks to the washing machine and sets it on top.
In S3, one hand opens and closes the air fryer drawer, while the other places the tomato inside.
\method thus keeps its advantage on a real humanoid, most clearly on the loco-manipulation tasks.

\noindent\textbf{Inference cost.}
The forecaster and engagement cross-attention layers bring \method to 3.45B parameters and 7.0\,GB of GPU memory, against 3.14B and 6.4\,GB for GR00T N1.7.
On an RTX 5080 workstation, its median prediction latency is 73\,ms and its 95th percentile is 81\,ms, against 65\,ms and 69\,ms for GR00T N1.7, while both models update action chunks about every 0.48\,s.

%% file: sec/table_intent_stats.tex
\caption{Median Pearson correlation over 720 episodes between the readout and engagement recovered from the executed commands. Contact compares Hands with simulator contact.}
\label{tab:intent-stats}
\centering
\footnotesize
\setlength{\tabcolsep}{4pt}
\begin{tabular*}{\linewidth}{@{\extracolsep{\fill}}lccccccc@{}}
  \toprule
   & Torso & Arms & Wrists & Hands & Base & All & Contact \\
  \midrule
  Readout & 0.89 & 0.92 & 0.75 & 0.98 & 0.89 & 0.92 & 0.92 \\
  Recovered & -- & -- & -- & -- & -- & -- & 0.96 \\
  \bottomrule
\end{tabular*}

%% file: sec/table_semisynth.tex
\caption{Engagement recovery on 260 semi-synthetic demonstrations. Arrows show true and recovered medians. Ramp is the rise before a switch, and lead is how early Arms rises before contact.}
\label{tab:semisynth}
\centering
\footnotesize
\begin{tabular*}{\linewidth}{@{\extracolsep{\fill}}lccc@{}}
  \toprule
  Truth & Ramp (s) & Correlation with truth & Lead (s) \\
  \midrule
  Gradual & 1.00 $\rightarrow$ 1.05 & 0.98 & 0.64 $\rightarrow$ 0.52 \\
  Abrupt & 0.02 $\rightarrow$ 0.56 & 0.92 & 0.36 $\rightarrow$ 0.24 \\
  \bottomrule
\end{tabular*}

%% file: sec/figure_real_results.tex
\begin{figure*}[t]
  \centering
  \includegraphics[width=\textwidth]{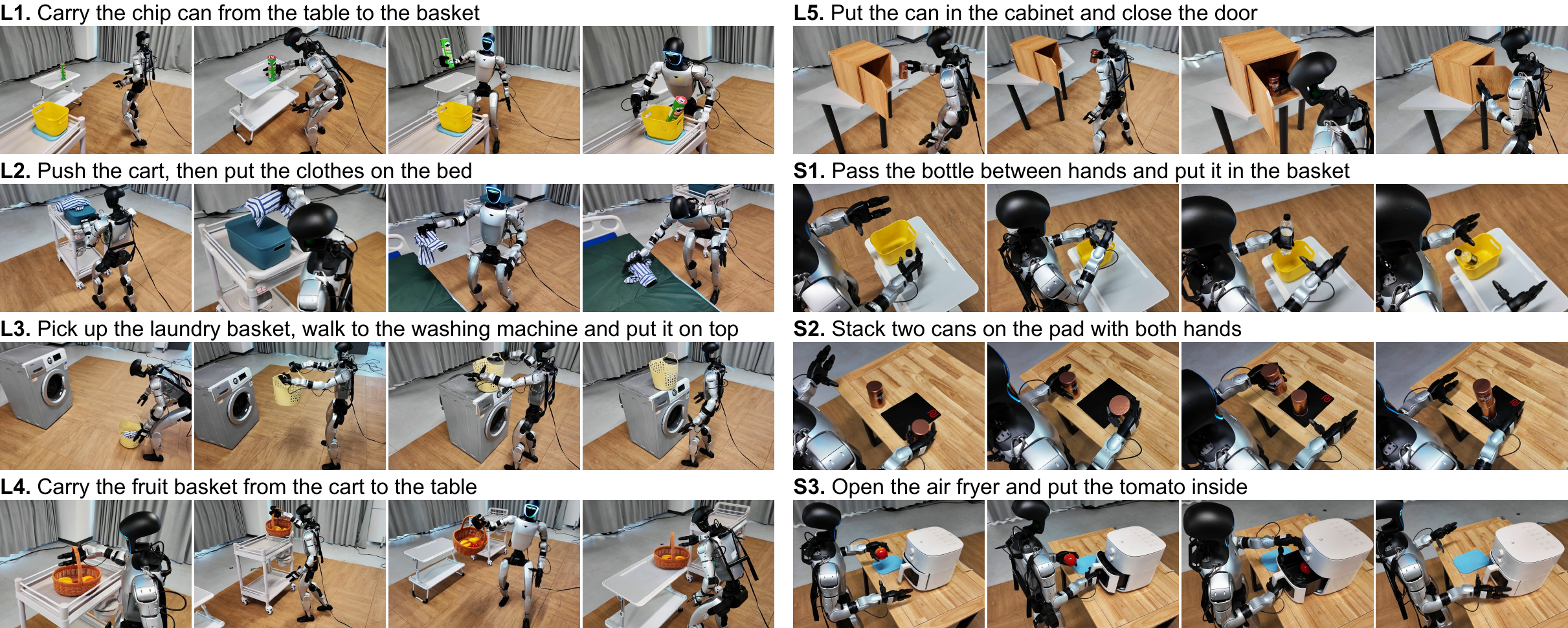}
  \caption{Real-world tasks on a Unitree G1, five loco-manipulation tasks L1 to L5 and three standing manipulation tasks S1 to S3 that need whole-body coordination without locomotion. Each task shows four keyframes of one autonomous run of \method, from left to right.}
  \label{fig:real-tasks}
  \captionof{table}{Real-world success rate (\%) over 20 trials per model and task on the tasks of \cref{fig:real-tasks}. Averages weight their tasks equally, the last three columns repeat L2, L3 and L4 with the object placed at random, and bold marks the highest value in each column.}
  \label{tab:real}
  \footnotesize
  \input{sec/table_real_body}
\end{figure*}

%% file: sec/table_real_body.tex
\ifdefined\realcolw\else\newlength{\realcolw}\newlength{\realavgw}\newlength{\realstandw}\newlength{\realrandw}\fi
\settowidth{\realcolw}{\textbf{00.0}}
\settowidth{\realavgw}{Standing}
\settowidth{\realstandw}{Standing manipulation}\divide\realstandw by 3
\settowidth{\realrandw}{Random placement}\divide\realrandw by 3
\ifdim\realstandw>\realcolw\setlength{\realcolw}{\realstandw}\fi
\ifdim\realrandw>\realcolw\setlength{\realcolw}{\realrandw}\fi
\setlength{\realstandw}{\realcolw}\setlength{\realrandw}{\realcolw}
\providecommand{\realcell}[1]{\makebox[\realcolw]{#1}}
\providecommand{\realavg}[1]{\makebox[\realavgw]{#1}}
\providecommand{\realstand}[1]{\makebox[\realstandw]{#1}}
\providecommand{\realrand}[1]{\makebox[\realrandw]{#1}}
\setlength{\tabcolsep}{0pt}
\begin{tabular*}{\textwidth}{@{\extracolsep{\fill}}lccccc@{\hspace{4pt}}ccc@{\hspace{4pt}}ccc@{\hspace{4pt}}ccc@{}}
  \toprule
  & \multicolumn{5}{c}{Loco-manipulation} & \multicolumn{3}{c}{Standing manipulation} & \multicolumn{3}{c}{Average} & \multicolumn{3}{c}{Random placement} \\
  \cmidrule(lr){2-6}\cmidrule(lr){7-9}\cmidrule(lr){10-12}\cmidrule(l){13-15}
  Model & \realcell{L1} & \realcell{L2} & \realcell{L3} & \realcell{L4} & \realcell{L5} & \realstand{S1} & \realstand{S2} & \realstand{S3} & \realavg{Loco} & \realavg{Standing} & \realavg{Overall} & \realrand{L2} & \realrand{L3} & \realrand{L4} \\
  \midrule
  GR00T N1.7 & \realcell{45.0} & \realcell{45.0} & \realcell{40.0} & \realcell{75.0} & \realcell{50.0} & \realstand{70.0} & \realstand{75.0} & \realstand{40.0} & \realavg{51.0} & \realavg{61.7} & \realavg{55.0} & \realrand{30.0} & \realrand{20.0} & \realrand{60.0} \\
  $\Psi_0$ & \realcell{55.0} & \realcell{30.0} & \realcell{45.0} & \realcell{80.0} & \realcell{40.0} & \realstand{60.0} & \realstand{\textbf{80.0}} & \realstand{40.0} & \realavg{50.0} & \realavg{60.0} & \realavg{53.8} & \realrand{25.0} & \realrand{25.0} & \realrand{50.0} \\
  \midrule
  \method{}, $\lambda = 0$ & \realcell{50.0} & \realcell{35.0} & \realcell{40.0} & \realcell{70.0} & \realcell{55.0} & \realstand{65.0} & \realstand{75.0} & \realstand{35.0} & \realavg{50.0} & \realavg{58.3} & \realavg{53.1} & \realrand{35.0} & \realrand{30.0} & \realrand{50.0} \\
  \method{} (ours) & \realcell{\textbf{75.0}} & \realcell{\textbf{65.0}} & \realcell{\textbf{60.0}} & \realcell{\textbf{85.0}} & \realcell{\textbf{75.0}} & \realstand{\textbf{75.0}} & \realstand{70.0} & \realstand{\textbf{50.0}} & \realavg{\textbf{72.0}} & \realavg{\textbf{65.0}} & \realavg{\textbf{69.4}} & \realrand{\textbf{45.0}} & \realrand{\textbf{50.0}} & \realrand{\textbf{85.0}} \\
  \bottomrule
\end{tabular*}

%% file: sec/6_conclusion.tex
\section{Conclusion}
\label{sec:conclusion}

We presented \method, which supervises an intermediate representation with body-part engagement to guide humanoid whole-body action generation.
Body-part engagement describes how several body parts engage together through target-directed motion and hand contact.
Engagement recovery anchors part targets to later events in teleoperation demonstrations and recovers target-directed engagement and hand contact without annotation.
The forecaster predicts body-part engagement over the action horizon from vision-language features and proprioception, and its latent representation conditions the action head.

\method achieves higher average success than four VLA and WAM baselines in simulation and higher overall success than the compared baselines on eight real-world tasks.
Eight-task ablations show gains over alternative supervision variants and the action head's reliance on the engagement representation.
Future work will incorporate visual scene context into engagement recovery using multimodal foundation models. We will build a large-scale body-part engagement dataset, pretrain the forecaster and evaluate transfer to new tasks involving different objects and scenes.